%% file: neurips_2025.tex
\documentclass{article}

    \PassOptionsToPackage{numbers, compress}{natbib}
\usepackage[preprint]{neurips_2025}

\usepackage{tikz}
\usepackage[utf8]{inputenc}
\usepackage[T1]{fontenc}
\usepackage[colorlinks=true]{hyperref}
\usepackage{url}
\usepackage{booktabs}
\usepackage{amsfonts}
\usepackage{amsmath}
\usepackage{amsthm}
\usepackage{amssymb}
\usepackage{nicefrac}
\usepackage{microtype}
\usepackage{xcolor}
\definecolor{mydeepblue}{RGB}{39, 60, 117}
\definecolor{myblue}{RGB}{200, 214, 229}
\hypersetup{linkcolor=mydeepblue,citecolor=mydeepblue,urlcolor=mydeepblue}  
\usepackage{graphicx}
\usepackage{subcaption}
\usepackage{multirow}
\usepackage{algorithm}
\usepackage{algorithmic}
\usepackage{xspace}
\usepackage{pifont}
\usepackage[framemethod=tikz]{mdframed}
\usepackage[most]{tcolorbox}

\newtheorem{theorem}{Theorem}

\newcommand{\method}{EVPO\xspace}
\newcommand{\ev}{\mathrm{ev}}
\newcommand{\Var}{\mathrm{Var}}

\newcommand{\E}{\mathbb{E}}

\usepackage{xspace}

\usepackage{cleveref}
\crefformat{section}{\S~#2#1#3}
\Crefformat{section}{\S~#2#1#3}
\crefname{section}{Appendix}{Appendices}
\Crefname{section}{Appendix}{Appendices}

\usepackage{booktabs}

\usepackage{booktabs}   
\usepackage{graphicx}   
\usepackage{multirow}   
\usepackage{amsmath}   
\usepackage[table]{xcolor} 

\usepackage{fontawesome5}

\newtcolorbox{titleEnv}{
colframe=black!80,
colback=gray!10,
fonttitle=\bfseries,
coltitle=black,
left=3pt,
right=3pt,
top=3pt,
bottom=3pt,
boxrule=0.4mm,
arc=2mm
}
\definecolor{mydeepblue}{RGB}{46, 90, 168}
\definecolor{myblue}{RGB}{166, 202, 236}
\definecolor{my_blue}{RGB}{0,120,255}
\definecolor{my_purple}{RGB}{161, 27, 155}
\definecolor{my_green}{RGB}{0, 176, 80}
\definecolor{msftBlue}{RGB}{0,164,239}
\definecolor{msftGreen}{RGB}{127,186,0}
\definecolor{msftYello}{RGB}{255,185,0}
\definecolor{msftBlack}{RGB}{0,0,0}

\newenvironment{findingBox}[2]{%
	\begin{tcolorbox}[
colframe=mydeepblue!80,
colback=myblue!50,
 boxrule=.5pt,
 left=1pt,
 right = 1pt,
 top= 0pt,
 bottom=0pt,
 size=small,
 fonttitle=\bfseries,
coltitle=black,
boxrule=0.4mm,
arc=2mm
 ]{\textbf{Finding #1:} #2} 
}{%
	\end{tcolorbox}
}

\usepackage{listings}

\lstdefinelanguage{json}{
  basicstyle=\ttfamily\scriptsize,
  showstringspaces=false,
  breaklines=true,
  columns=fullflexible,
  morestring=[b]",
  morekeywords={true,false,null}
}

\title{EVPO: Explained Variance Policy Optimization for\\Adaptive Critic Utilization in LLM Post-Training}

\author{
    Chengjun Pan\textsuperscript{1}\thanks{Equal contribution. $^\dagger$Corresponding authors:\texttt{tgui@fudan.edu.cn, fengyansong@pku.edu.cn}},
    Shichun Liu\textsuperscript{2}$^*$,
    Jiahang Lin\textsuperscript{2}$^*$,
    Dingwei Zhu\textsuperscript{2},\\ 
    \textbf{Jiazheng Zhang\textsuperscript{2},
    Shihan Dou\textsuperscript{2},
    Songyang Gao\textsuperscript{4},
    Zhenhua Han\textsuperscript{3},}
    \\
    \textbf{Binghai Wang\textsuperscript{2},
    Rui Zheng\textsuperscript{3},
    Xuanjing Huang\textsuperscript{2},
    Tao Gui\textsuperscript{2}$^\dagger$,
    Yansong Feng\textsuperscript{1}$^\dagger$}
    \\
    \\
  \textsuperscript{1}Peking University,
  \textsuperscript{2}Fudan University,
  \textsuperscript{3}Shanghai Qiji Zhifeng Co., Ltd.,
  \textsuperscript{4}Shanghai AI Lab
}

\begin{document}

\maketitle

\begin{abstract}
Reinforcement learning (RL) for LLM post-training faces a fundamental design choice: whether to use a learned critic as a baseline for policy optimization. Classical theory favors critic-based methods such as PPO for variance reduction, yet critic-free alternatives like GRPO have gained widespread adoption due to their simplicity and competitive performance.
We show that in sparse-reward settings, a learned critic can inject estimation noise that exceeds the state signal it captures, increasing rather than reducing advantage variance.
By casting baseline selection as a Kalman filtering problem, we unify PPO and GRPO as two extremes of the Kalman gain and prove that \emph{explained variance} (EV), computable from a single training batch, identifies the exact boundary: positive EV indicates the critic reduces variance, while zero or negative EV signals that it inflates variance.
Building on this insight, we propose \textbf{E}xplained \textbf{V}ariance \textbf{P}olicy \textbf{O}ptimization (\textbf{EVPO}), which monitors batch-level EV at each training step and adaptively switches between critic-based and batch-mean advantage estimation, provably achieving no greater variance than the better of the two at every step.
Across four tasks spanning classical control, agentic interaction, and mathematical reasoning, \method consistently outperforms both PPO and GRPO regardless of which fixed baseline is stronger on a given task.
Further analysis confirms that the adaptive gating tracks critic maturation over training and that the theoretically derived zero threshold is empirically optimal.
\end{abstract}

\begin{figure}[htbp]
    \centering
    \includegraphics[width=0.9\textwidth]{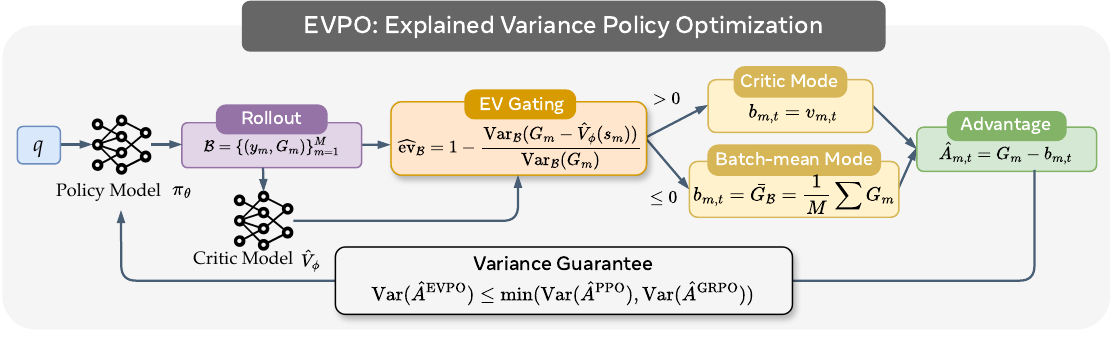}
    \caption{Overview of \method. The EV gating module selects between the critic baseline (EV $> 0$) and the batch-mean baseline (EV $\leq 0$), guaranteeing advantage variance no greater than the minimum of PPO and GRPO.}
    \label{fig:evpo-main}
\end{figure}

\input{sections/Introduction}

\input{sections/Preliminary}

\input{sections/Method}

\input{sections/Experiments}

\input{sections/Related_Work}

\input{sections/Conclusion}

\input{sections/Limitations}

\bibliographystyle{plainnat}
\bibliography{references}

\appendix

\input{sections/app_preliminary}
\input{sections/app_exp_details}

\end{document}

%% file: sections/Introduction.tex
\section{Introduction}
\label{sec:intro}

Reinforcement learning (RL) has become a central paradigm for LLM post-training, driving advances in mathematical reasoning~\citep{guoDeepSeekR1IncentivizesReasoning2025,singhOpenAIGPT5System2025,qwen3.5_toward_native,yangQwen3TechnicalReport2025}, instruction following~\citep{zhouInstructionFollowingEvaluationLarge2023a,dingOctoBenchBenchmarkingScaffoldAware2026,douCLbenchBenchmarkContext2026}, and interactive decision-making~\citep{shridharALFWorldAligningText2021,wangRAGENUnderstandingSelfEvolution2025,chenGUIShepherdReliableProcess2025}.
A prominent instantiation is \emph{reinforcement learning with verifiable rewards} (RLVR)~\citep{guoDeepSeekR1IncentivizesReasoning2025}, where a sparse reward is assigned only at the terminal step via an automatic verifier.
A fundamental algorithmic choice in this setting is whether to employ a \emph{critic} (a learned value function) as a baseline for policy optimization.

Classical RL theory holds that a critic reduces policy-gradient variance by providing a state-dependent baseline~\citep{suttonPolicyGradientMethods1999,schulmanHighDimensionalContinuousControl2018}, forming the foundation of actor-critic methods including PPO~\citep{schulmanProximalPolicyOptimization2017}.
In practice, however, critic-free methods such as GRPO~\citep{shaoDeepSeekMathPushingLimits2024,guoDeepSeekR1IncentivizesReasoning2025} have gained widespread adoption, as they are cheaper and can match or even surpass PPO~\citep{huREINFORCEStabilizingCriticFree2025}.
Rather than treating the critic as a binary design choice, we ask:
\begin{center}
\emph{In sparse-reward LLM post-training, when does a learned critic help and when does it hurt?}
\end{center}
To answer this, we identify three progressive sub-questions.
{\color{purple}$\mathcal{Q}_1$:~\emph{Is the boundary observable?}} Critic quality fluctuates during training, and we need to measure its impact on policy optimization in a controlled setting.
{\color{brown}$\mathcal{Q}_2$:~\emph{Where exactly is the boundary?}} In sparse-reward LLM generation, the critic may inject more estimation noise than the state signal it captures. We need a precise characterization of when this happens.
{\color{teal}$\mathcal{Q}_3$:~\emph{Can this boundary be exploited online?}} If the helpful-vs-harmful regime can be detected in real time, we can build an algorithm that adapts rather than committing to a fixed choice.

We adopt \emph{explained variance} (EV), the fraction of return variance accounted for by the critic, as a real-time proxy for critic usefulness.
Empirically ({\color{purple}$\mathcal{Q}_1$}), EV remains negative during the early stage of training, and controlled interventions (suppressing the critic or injecting noise) confirm that performance degrades once EV falls to zero or below, suggesting that its sign reliably indicates whether the critic helps or hurts (Section~\ref{ssec:observations}).
To pinpoint the boundary ({\color{brown}$\mathcal{Q}_2$}), we cast baseline selection as a Kalman filtering problem~\citep{kalmanNewApproachLinear1960} in which the critic and the batch-mean return are two competing estimators of the true state value, with PPO and GRPO corresponding to the two extremes of the Kalman gain (Section~\ref{sec:kalman}).
We prove that the sign of EV, computable from a single batch, is sufficient to determine which extreme is preferable (\Cref{thm:collapse}), placing the boundary at $\text{EV} = 0$.

Exploiting this boundary online ({\color{teal}$\mathcal{Q}_3$}), we propose \textbf{Explained Variance Policy Optimization (\method)}, which implements this adaptive hard gating at each training step.
\method maintains a standard PPO-style critic alongside the policy, and at each iteration computes batch-level EV as a scalar indicator.
When EV is positive, the critic captures meaningful state-dependent signal, and \method uses its predictions as the baseline (the PPO mode);
when EV falls to zero or below, \method discards the critic and falls back to the batch-mean return (the GRPO mode).
We prove that this switching rule yields advantage-estimator variance no greater than the better of PPO and GRPO at every step, so \method never underperforms either fixed choice.



We evaluate \method on four tasks spanning classical control, interactive decision-making, and mathematical reasoning.
Across all tasks, \method achieves the best final performance and maintains its advantage throughout training, regardless of whether PPO or GRPO is the stronger fixed baseline on a given task.
Further analysis confirms that the adaptive gating tracks critic maturation over training and that the theoretically motivated zero threshold is empirically optimal.

Our main contributions are:
\begin{enumerate}
    \item \textbf{Indicator and theory.} We unify PPO and GRPO as two extremes of a Kalman baseline-selection spectrum and prove that the sign of explained variance is the exact boundary separating the variance-reducing from the variance-inflating critic regime (Theorem~\ref{thm:collapse}).
    \item \textbf{Algorithm.} 
    We propose \method, a lightweight method that switches between critic-based and critic-free advantage estimation via batch-level EV, with a provable guarantee of no greater variance than the better of PPO and GRPO at every step.
    \item \textbf{Empirical validation.} \method consistently outperforms fixed PPO and GRPO across four tasks spanning classical control, agentic interaction, and mathematical reasoning. Analysis experiments confirm that the zero threshold is empirically optimal and that the adaptive gating reflects critic maturation across training stages.
\end{enumerate}

%% file: sections/Preliminary.tex
\section{Preliminaries}
\label{sec:prelim}

This section establishes that the critic is not uniformly beneficial. We review PPO and GRPO as two baseline-selection extremes (Section~\ref{ssec:rl_for_llm_post_training}), then show empirically that the sign of explained variance reliably predicts whether the critic helps or hurts (Section~\ref{ssec:observations}).

\subsection{RL for LLM Post-Training}
\label{ssec:rl_for_llm_post_training}

We model autoregressive text generation as a token-level Markov Decision Process~\citep{openaiOpenAIO1System2024,trungReFTReasoningReinforced2024,zhengSecretsRLHFLarge2023}.
Given a prompt $x$, the policy $\pi_\theta$ generates a response $y = (y_1, \ldots, y_T)$ token by token, where the state at step $t$ is $s_t = (x, y_{<t})$ and the action is $a_t = y_t \sim \pi_\theta(\cdot \mid s_t)$.

In the reinforcement learning with verifiable rewards (RLVR) setting~\citep{guoDeepSeekR1IncentivizesReasoning2025,wangRAGENUnderstandingSelfEvolution2025}, a scalar reward is assigned only at the terminal step: $r_T = \mathcal{R}(x, y)$, $r_t = 0$ for $t < T$.
With discount factor $\gamma = 1$, the return from any state equals the terminal reward: $G_t = r_T \triangleq G$.
The state-value function is $V^{\pi}(s) = \E_{\pi}[G \mid s_t = s]$, and the observed return $G$ is a stochastic realization whose variance arises solely from the randomness of future actions under $\pi$.

\paragraph{PPO and GRPO.}
Both PPO~\citep{schulmanProximalPolicyOptimization2017} and GRPO~\citep{guoDeepSeekR1IncentivizesReasoning2025,shaoDeepSeekMathPushingLimits2024} optimize a clipped surrogate objective but differ in how they estimate advantages.
Under RLVR conditions ($\gamma{=}\lambda{=}1$, terminal-only reward), the PPO advantage reduces to (see Appendix~\ref{app:gae_derivation} for the full GAE derivation):
\begin{equation}
  \hat{A}_t^{\mathrm{PPO}} = G - \hat{V}_\phi(s_t),
  \label{eq:ppo_adv}
\end{equation}
where $\hat{V}_\phi$ is a learned critic providing a \emph{state-dependent} baseline.
GRPO eliminates the critic and instead computes a \emph{response-level} advantage from batch statistics.
Concretely, for each prompt $x$, a batch of $K$ responses $\{y^{(i)}\}_{i=1}^{K}$ is sampled from the current policy, yielding returns $\{G_i\}_{i=1}^{K}$ with batch mean $\bar{G}_{\mathcal{B}} = \frac{1}{K}\sum_{i=1}^{K} G_i$ and standard deviation $\sigma(G_{\mathcal{B}})$.
The GRPO advantage is then:
\begin{equation}
  \hat{A}_i^{\mathrm{GRPO}} = \frac{G_i - \bar{G}_{\mathcal{B}}}{\sigma(G_{\mathcal{B}})},
  \label{eq:grpo_adv}
\end{equation}
where $\bar{G}_{\mathcal{B}}$ and $\sigma(G_{\mathcal{B}})$ are the batch mean and standard deviation of returns, serving as a \emph{constant} baseline shared across all tokens.


\subsection{The Impact of Critic Quality on Policy Optimization}
\label{ssec:observations}

Returning to {\color{purple}$\mathcal{Q}_1$} (is the boundary observable?), we quantify critic quality via the \emph{explained variance} (EV) metric, a standard indicator from regression analysis~\citep{schulmanNutsBoltsDeep}:
\begin{equation}
  \ev = 1 - \frac{\Var(G - \hat{V}_\phi(s))}{\Var(G)},
  \label{eq:ev_def}
\end{equation}
which measures the fraction of return variance accounted for by the critic.
Intuitively, $\ev > 0$ indicates that the critic captures meaningful state-dependent signal, whereas $\ev \le 0$ means the critic's estimation noise exceeds the signal it provides---its predictions are worse than a constant baseline.

\begin{figure}[htbp]
    \centering
        \begin{subfigure}[b]{0.24\textwidth}
            \centering
            \includegraphics[width=\textwidth]{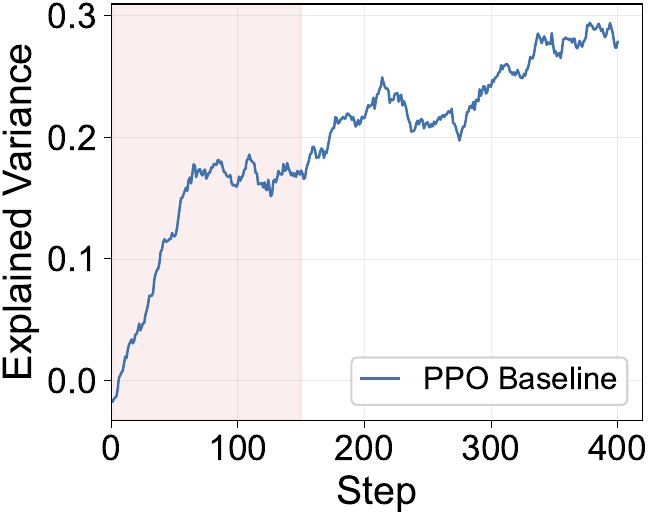}
            \caption{Explained Variance}
            \label{fig:obs1a}
        \end{subfigure}
        \hfill
        \begin{subfigure}[b]{0.24\textwidth}
            \centering
            \includegraphics[width=\textwidth]{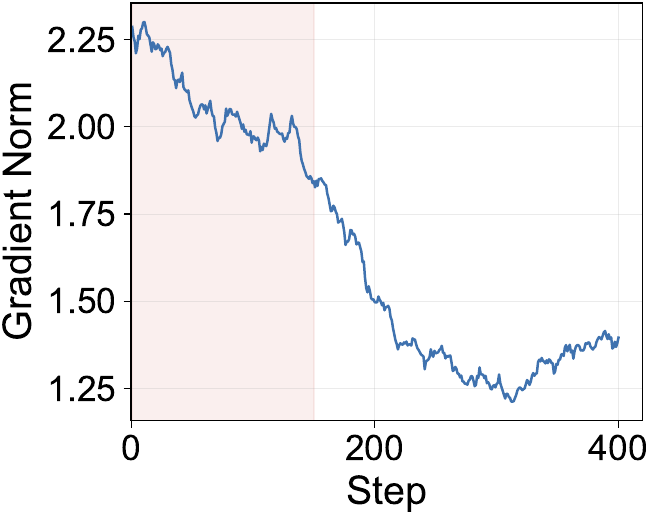}
            \caption{Gradient Norm}
            \label{fig:obs1b}
        \end{subfigure}
        \hfill
        \begin{subfigure}[b]{0.24\textwidth}
            \centering
            \includegraphics[width=\textwidth]{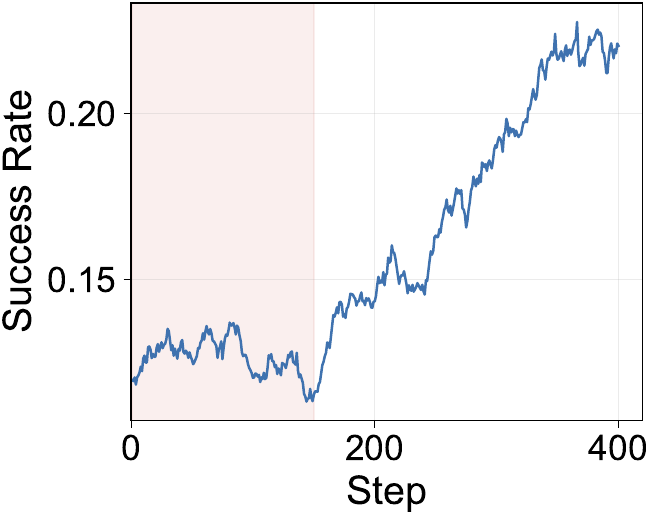}
            \caption{Train Success}
            \label{fig:obs1c}
        \end{subfigure}
        \hfill
        \begin{subfigure}[b]{0.24\textwidth}
            \centering
            \includegraphics[width=\textwidth]{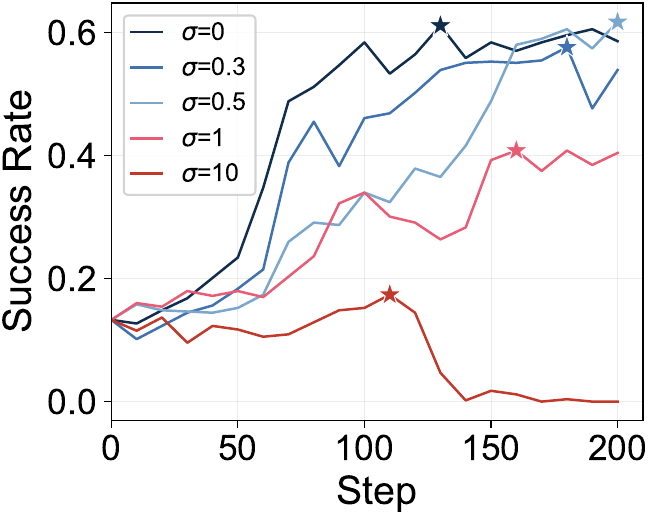}
            \caption{Noise Injection}
            \label{fig:obs1d}
        \end{subfigure}
    \caption{Training dynamics of PPO on Sokoban. \textbf{(a)}~Batch-level EV over training steps. \textbf{(b)}~Policy gradient norms. \textbf{(c)}~Training success rate. The shaded region in (a--c) marks the early phase where EV is low. \textbf{(d)}~Effect of injecting Gaussian noise into a converged critic on FrozenLake.}
    \label{fig:obs1}
\end{figure}

\paragraph{Observation 1: A noisy critic hurts the actor.}
We train a PPO policy on Sokoban and track the batch-level EV throughout training (see Appendix~\ref{app:preliminary_exp1} for details).
As shown in Figure~\ref{fig:obs1a}, EV remains consistently low during the first ${\sim}150$ training steps, indicating that the critic's predictions are less accurate than a simple constant baseline during this phase.
This poor critic quality has two visible consequences: gradient norms are notably elevated (Figure~\ref{fig:obs1b}), reflecting high-variance policy updates driven by noisy advantage estimates, and the training success rate remains low and unstable (Figure~\ref{fig:obs1c}), showing that the actor struggles to learn under unreliable critic guidance.

To further confirm the causal link between critic quality and actor performance, we conduct a controlled experiment: injecting Gaussian noise ($\sigma_\delta = 0.3$; the mean absolute state value is $\approx 0.5$) into a \emph{converged} critic's outputs.
As shown in Figure~\ref{fig:obs1d}, performance degrades monotonically as the noise level increases, with complete collapse at $\sigma{=}10$ where EV is driven far below zero.
This confirms that \textbf{negative EV is a reliable indicator of harmful critic influence}, and that the relationship between critic quality and actor performance is not merely correlational but causal.

\begin{figure}[ht]
  \centering
      \begin{subfigure}[b]{0.48\textwidth}
          \centering
          \includegraphics[width=0.49\textwidth]{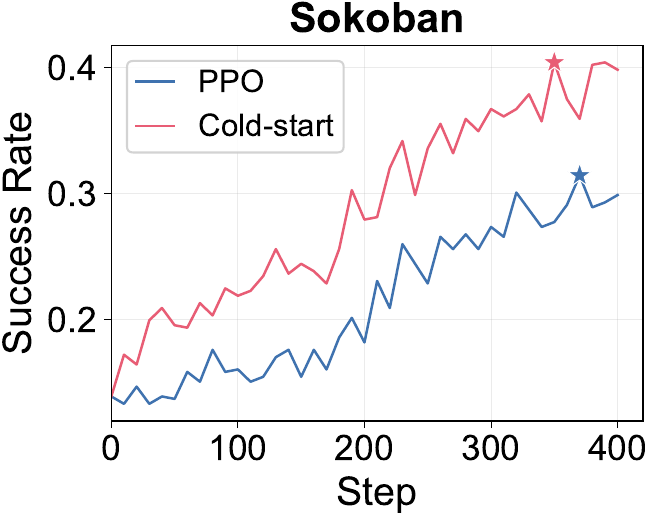}
          \hfill
          \includegraphics[width=0.49\textwidth]{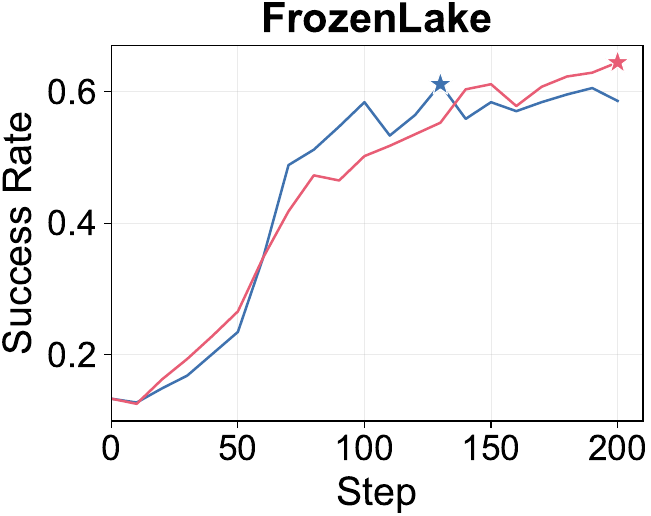}
          \caption{Cold-start}
      \end{subfigure}
      \hfill
      \begin{subfigure}[b]{0.48\textwidth}
          \centering
          \includegraphics[width=0.49\textwidth]{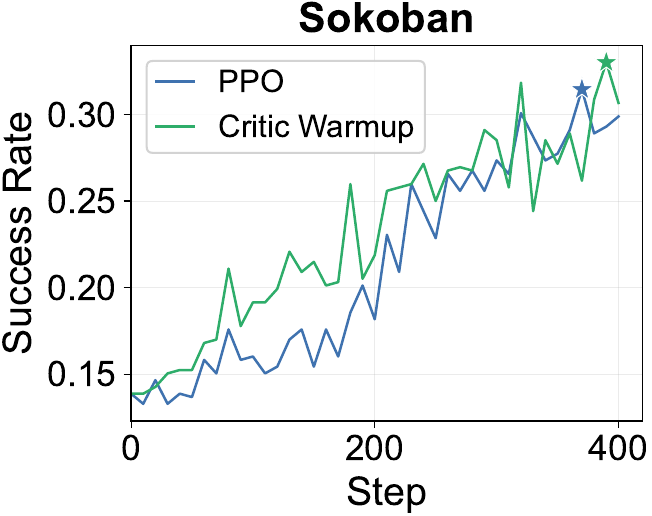}
          \hfill
          \includegraphics[width=0.49\textwidth]{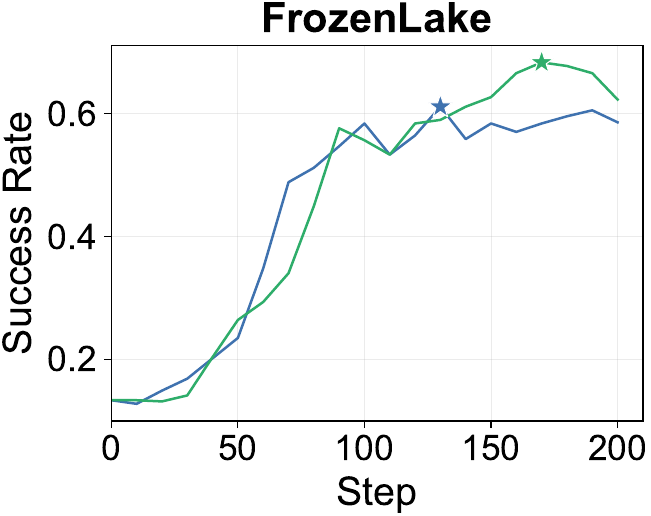}
          \caption{Critic Warmup}
      \end{subfigure}
    \caption{Reducing critic variance helps the actor. \textbf{(a)}~Cold-start: replacing the critic output with $\bar{G}_{\mathcal{B}}$ during early training on Sokoban (left) and FrozenLake (right). \textbf{(b)}~Critic Warmup: pre-training the critic before actor training begins.}
    \label{fig:obs2}
    \end{figure}
    
\paragraph{Observation 2: Reducing critic variance helps the actor.}
We conduct two complementary interventions on the same Sokoban and FrozenLake setups (see Appendix~\ref{app:preliminary_exp2} for details):
(i)~\emph{Cold-start}: replacing the critic's output with the batch mean $\bar{G}_{\mathcal{B}}$ during the first 50/25 steps, and
(ii)~\emph{Warm-up}: pre-training the critic alone for 200 steps before the actor begins updating.
Both interventions yield faster convergence and more stable training curves (Figure~\ref{fig:obs2}), confirming that \textbf{suppressing critic variance during low-quality phases directly benefits the policy}.

These observations suggest that the choice between a learned critic and a batch-mean baseline should not be fixed a priori, but should adapt to the critic's current quality.
In the following subsection, we formalize this intuition through a Kalman filtering lens.

\subsection{A Kalman Filtering Perspective on Baseline Selection}
\label{sec:kalman}

To precisely locate the boundary ({\color{brown}$\mathcal{Q}_2$}), we observe that choosing a baseline is fundamentally a \emph{state-estimation} problem: both the critic $\hat{V}_\phi(s)$ and the batch mean $\bar{G}_{\mathcal{B}}$ are noisy estimates of the true value $V^\pi(s)$, and the best baseline is the one with lower estimation error.
This is exactly the setting addressed by the Kalman filter~\citep{kalmanNewApproachLinear1960,peiElementaryIntroductionKalman2019}, which optimally fuses two noisy estimates by weighting each inversely to its error variance.

\paragraph{Setup.}
The observed return decomposes as $G = V^\pi(s) + \epsilon$ with sampling noise $R \triangleq \E[\Var(G \mid s)]$.
The two competing baselines have distinct error structures: (i)~the \emph{critic} $\hat{V}_\phi(s) = V^\pi(s) + \delta(s)$ with estimation error variance $P_A \triangleq \Var(\delta)$, and (ii)~the \emph{batch mean} $\bar{G}_{\mathcal{B}}$ whose deviation from any specific state value has variance $P_B \triangleq \Var_s(V^\pi(s))$.
The minimum-variance linear combination of these two estimators assigns the critic a weight of $1 - K^*$ with the optimal gain $K^* = P_A / (P_A + P_B)$ (see Appendix~\ref{app:kalman_details} for the full derivation).

\paragraph{Connection to explained variance.}
The EV metric (Eq.~\eqref{eq:ev_def}) directly encodes the relative reliability of the two estimators.
Substituting the variance decomposition into the definition of EV yields:
\begin{equation}
  \ev = \frac{P_B - P_A}{P_B + R}.
  \label{eq:ev_kalman}
\end{equation}
Since the denominator $P_B + R$ is always positive, the \emph{sign} of EV is determined solely by $P_B - P_A$.
This means \textbf{$\ev = 0$ marks the exact boundary where $P_A = P_B$}, i.e., where the critic's estimation noise equals the state signal it captures.
Above this boundary ($\ev > 0$), the critic outperforms the batch mean; below it ($\ev \le 0$), critic noise dominates.
The sampling noise $R$ compresses the magnitude of EV but preserves its sign, keeping this boundary robust under high stochasticity.


%% file: sections/Method.tex
\section{Methodology}
\label{sec:method}

Building on the Kalman filtering framework of Section~\ref{sec:kalman}, we now formalize when a critic helps or hurts (Section~\ref{sec:ev_threshold}), unify PPO and GRPO as two extremes of the Kalman gain (Section~\ref{sec:unification}), and derive the \method algorithm (Section~\ref{sec:evpo}).

\subsection{The Performance Collapse Boundary}
\label{sec:ev_threshold}

From Eq.~\eqref{eq:ev_kalman}, the sign of EV is governed by $P_B - P_A$. We formalize this as:

\begin{tcolorbox}[
colframe=mydeepblue!80,
colback=myblue!50,
boxrule=.5pt,
left=1pt,
right = 1pt,
top= 0pt,
bottom=0pt,
size=small,
fonttitle=\bfseries,
coltitle=black,
boxrule=0.4mm,
arc=2mm
]
  \begin{theorem}[Performance Collapse Boundary]
  \label{thm:collapse}
  Let $\hat{V}_\phi(s) = V^\pi(s) + \delta(s)$ with $\delta \perp \epsilon$, and let $P_A = \Var(\delta)$, $P_B = \Var_s(V^\pi(s))$, $K = P_A/(P_A+P_B)$.
  Then the following are equivalent:
  \begin{equation}
    \ev \le 0 \;\;\Longleftrightarrow\;\; P_A \ge P_B \;\;\Longleftrightarrow\;\; K \ge \tfrac{1}{2}.
    \label{eq:collapse}
  \end{equation}
  \end{theorem}
\end{tcolorbox}
\noindent
In words, the critic inflates advantage variance whenever its estimation noise ($P_A$) exceeds the state signal it captures ($P_B$), and $\ev = 0$ is the exact tipping point. Since the sign of EV is computable from a single batch, this boundary is directly observable during training.

\paragraph{Computing EV from a training batch.}
Replacing the population variances in Eq.~\eqref{eq:ev_def} with sample variances over a batch of $M$ rollouts $\mathcal{B} = \{(y_m, G_m)\}_{m=1}^M$ from the same prompt gives the batch-level EV:
\begin{equation}
  \widehat{\ev}_{\mathcal{B}} = 1 - \frac{\mathrm{Var}_{\mathcal{B}}(G_m - \hat{V}_\phi(s_m))}{\mathrm{Var}_{\mathcal{B}}(G_m)},
  \label{eq:ev_batch}
\end{equation}
computed from the critic values already obtained for advantage estimation, with no additional forward passes.

\subsection{PPO and GRPO as Kalman Extremes}
\label{sec:unification}

Under RLVR conditions ($\gamma{=}\lambda{=}1$, terminal-only reward), PPO and GRPO can be unified into the form $\hat{A}_t = G - b(s_t)$, differing only in the choice of baseline $b(s_t)$ (see Appendix~\ref{app:gae_derivation} for the detailed derivation).
This difference maps directly onto the Kalman gain $K$:

\begin{itemize}
  \item \textbf{PPO} (critic mode, $K{=}0$, full critic trust): $b(s_t) = \hat{V}_\phi(s_t)$. The baseline is entirely model-driven; the Kalman filter places zero weight on the evidence-driven estimator.
  \item \textbf{GRPO} (batch-mean mode, $K{=}1$, no critic trust): $b(s_t) = \bar{G}_{\mathcal{B}}$. (Since $1/\sigma(G_{\mathcal{B}})$ does not affect the gradient direction.) The baseline is entirely evidence-driven; the critic is discarded and the batch mean serves as a constant baseline for all tokens.
\end{itemize}

\noindent
In other words, PPO and GRPO are not fundamentally different algorithms but rather two extreme operating points of the same Kalman baseline-selection spectrum.
PPO implicitly assumes $P_A \ll P_B$ (the critic is always reliable), while GRPO implicitly assumes $P_A \gg P_B$ (the critic is never worth using).
Neither assumption holds uniformly across training, motivating an adaptive approach.

\begin{algorithm}[t]
  \caption{Explained Variance Policy Optimization (\method)}
  \label{alg:evpo}
  \begin{algorithmic}[1]
  \REQUIRE Policy $\pi_\theta$, critic $\hat{V}_\phi$, task set $\mathcal{D}$, group size $M$, number of tasks $N$
  \FOR{each training iteration}
    \STATE Sample $N$ tasks $\{x_j\}_{j=1}^N \sim \mathcal{D}$
    \FOR{$j = 1, \ldots, N$}
      \STATE Rollout a batch with $M$ trajectories $\mathcal{B}=\{y_m\}_{m=1}^M \sim \pi_\theta(\cdot \mid x_j)$
      \STATE Obtain rewards $G_m = \mathcal{R}(x_j, y_m)$; compute critic values $v_{m,t} \leftarrow \hat{V}_\phi(s_{m,t})$
      \STATE $\widehat{\ev}_{\mathcal{B}} \leftarrow 1 - \mathrm{Var}_{\mathcal{B}}(G_m - v_m) \;/\; \mathrm{Var}_{\mathcal{B}}(G_m)$
      \IF{$\widehat{\ev}_{\mathcal{B}} > 0$}
        \STATE $b_{m,t} \leftarrow v_{m,t}$ \hfill $\triangleright$ \textit{Critic mode}
      \ELSE
        \STATE $b_{m,t} \leftarrow \bar{G}_{\mathcal{B}} = \frac{1}{M}\sum_{m=1}^{M} G_m$ \hfill $\triangleright$ \textit{Batch-mean mode}
      \ENDIF
      \STATE $\hat{A}_{m,t} \leftarrow G_m - b_{m,t}$
    \ENDFOR
    \STATE Aggregate gradients over $N$ tasks; update $\pi_\theta$ and $\hat{V}_\phi$
  \ENDFOR
  \end{algorithmic}
\end{algorithm}

\subsection{EVPO: Adaptive Baseline Selection via EV Gating}
\label{sec:evpo}

Having established \emph{where} the collapse boundary lies ({\color{brown}$\mathcal{Q}_2$}), we now turn to {\color{teal}$\mathcal{Q}_3$}: \emph{can this boundary be exploited online?}
The Kalman framework prescribes the optimal baseline as a continuous interpolation of the two estimators:
\begin{equation}
  b(s) = (1 - K)\,\hat{V}_\phi(s) + K\,\bar{G}_{\mathcal{B}}, \quad K = \frac{P_A}{P_A + P_B} \in [0,1].
  \label{eq:kalman_baseline}
\end{equation}
However, computing $K$ directly is infeasible: both $P_A = \Var(\hat{V}_\phi(s) - V^\pi(s))$ and $P_B = \Var_s(V^\pi(s))$ depend on the unknown true value $V^\pi(s)$, making the continuous $K$ non-identifiable in practice.
We therefore reduce the domain of $K$ to $\{0,1\}$, corresponding to the two extremes: $K{=}0$ (PPO, full critic trust) and $K{=}1$ (GRPO, no critic).
The decision boundary $\frac{P_A}{P_A+P_B} = \tfrac{1}{2}$ (i.e., $P_A = P_B$) corresponds exactly to $\ev = 0$ by Theorem~\ref{thm:collapse}.
While the \emph{value} of $K$ is non-identifiable, its position relative to $\tfrac{1}{2}$ can be determined by the sign of $\widehat{\ev}_{\mathcal{B}}$ (Eq.~\eqref{eq:ev_batch}).
This yields the \textbf{\method switching rule}:
\begin{equation}
  b(s_t) =
  \begin{cases}
    \hat{V}_\phi(s_t), & \text{if } \widehat{\ev}_{\mathcal{B}} > 0 \quad (K < \tfrac{1}{2}:\;\text{trust critic}), \\[4pt]
    \bar{G}_{\mathcal{B}},       & \text{if } \widehat{\ev}_{\mathcal{B}} \le 0 \quad (K \ge \tfrac{1}{2}:\;\text{retreat to batch mean}),
  \end{cases}
  \label{eq:evpo_rule}
\end{equation}
where $\widehat{\ev}_{\mathcal{B}}$ is the batch-level EV defined in Eq.~\eqref{eq:ev_batch}.

\paragraph{Variance guarantee.}
By construction, EVPO selects the lower-variance baseline at each step:
\begin{equation}
  \Var(\hat{A}^{\mathrm{EVPO}}) = R + \min(P_A,\; P_B),
  \label{eq:evpo_var}
\end{equation}
which is no greater than $\Var(\hat{A}^{\mathrm{PPO}}) = R + P_A$ or $\Var(\hat{A}^{\mathrm{GRPO}}) = R + P_B$ at every training step.
Over the course of training, this per-step guarantee accumulates: EVPO avoids the critic's noisy cold-start phase ($P_A \gg P_B$) while fully exploiting a mature critic ($P_A \ll P_B$).

%% file: sections/Experiments.tex
\section{Experiments}
\label{sec:experiments}

Our experiments aim to validate that the per-step variance reduction of \method (Eq.~\ref{eq:evpo_var}) translates into consistent performance gains over fixed baselines (Section~\ref{sec:main_results}). We further analyze whether EV serves as a reliable indicator of the $P_A$-vs-$P_B$ balance throughout training (Section~\ref{sec:gating}) and whether the theoretically derived collapse boundary $\ev = 0$ (Theorem~\ref{thm:collapse}) coincides with the empirically optimal switching threshold (Section~\ref{sec:threshold}).

\subsection{Experimental Setup}
\label{sec:exp_setup}
\paragraph{Evaluation.}
We evaluate on four tasks that span diverse agent interaction paradigms. \textbf{Sokoban}~\citep{junghannsSokobanEnhancingGeneral2001} and \textbf{FrozenLake}~\citep{PDFFrozenLake} are multi-turn grid-world games requiring sequential planning under sparse rewards, with FrozenLake additionally introducing environmental stochasticity. \textbf{WebShop}~\citep{yaoWebShopScalableRealWorld2022} is an interactive web-based agentic task involving multi-step instruction following and decision making. \textbf{MATH} targets competition-level mathematical reasoning trained and evaluated on problems in DAPO-Math-17k~\citep{yuDAPOOpenSourceLLM2025a}. 

\paragraph{Baseline.}
We compare \method{} against the \textbf{Base LLM} without RL post-training, two critic-based methods (\textbf{PPO}~\citep{schulmanProximalPolicyOptimization2017} and \textbf{StarPO-S}~\citep{wangRAGENUnderstandingSelfEvolution2025}), and two critic-free methods (\textbf{GRPO}~\citep{shaoDeepSeekMathPushingLimits2024} and \textbf{DAPO}~\citep{yuDAPOOpenSourceLLM2025a}). For the Math task we use Qwen2.5-7B-Instruct~\citep{qwenQwen25TechnicalReport2025} as the base model; the other three tasks use Qwen2.5-3B-Instruct~\citep{qwenQwen25TechnicalReport2025}. To ensure fair comparison, the hyperparameter recipe is tuned independently per task but kept consistent across all methods on the same task. Full configurations are provided in Appendix~\ref{app:exp_details}.

\subsection{Main Results}
\label{sec:main_results}

\begin{table}[htbp]
    \centering
    \caption{Best validation success rate across four tasks. \textbf{Bold} indicates the best result per task. For each task, the success rate is calculated based on a total of 512 experimental results, obtained by sampling 16 times for each of the 32 queries.}
    \label{tab:main}
    \vspace{0.5em}
    
    \resizebox{0.7\textwidth}{!}{%
    \begin{tabular}{lcccc}
    \toprule
    \textbf{Method} & \textbf{Sokoban} & \textbf{FrozenLake} & \textbf{WebShop} & \textbf{MATH} \\
    
    \midrule
    Base LLM & 0.139 & 0.133 & 0.025 & 0.332 \\
    \midrule
    \rowcolor{gray!15} \multicolumn{5}{l}{\textit{Critic-based methods}} \\
    PPO & 0.314 & 0.611 & 0.252 & 0.373 \\
    StarPO-S & 0.455 & 0.359 & 0.178 & 0.371 \\
    \midrule
    \rowcolor{gray!15} \multicolumn{5}{l}{\textit{Critic-free methods}} \\
    GRPO & 0.156 & 0.234 & 0.143 & 0.379 \\
    DAPO & 0.162 & 0.207 & 0.127 & 0.385 \\
    \midrule
    \textbf{\method}(\textit{ours}) & \textbf{0.604} & \textbf{0.684} & \textbf{0.303} & \textbf{0.416} \\
    \bottomrule
    \end{tabular}
    }

\end{table}

\begin{findingBox}{1}{
    \method improves over both PPO and GRPO regardless of which fixed baseline is stronger on a given task.
}
\end{findingBox}
The stronger fixed baseline varies across tasks: critic-based methods (PPO, StarPO-S) clearly outperform critic-free methods (GRPO, DAPO) on Sokoban, FrozenLake, and WebShop, while critic-free methods are more competitive on MATH. \method surpasses both families in every case (Table~\ref{tab:main}), confirming that it is not tied to a single favorable regime but adapts to the task-specific $P_A$-vs-$P_B$ balance.

\begin{figure}[htbp]
\centering
    \begin{subfigure}[b]{0.24\textwidth}
        \centering
        \includegraphics[width=\textwidth]{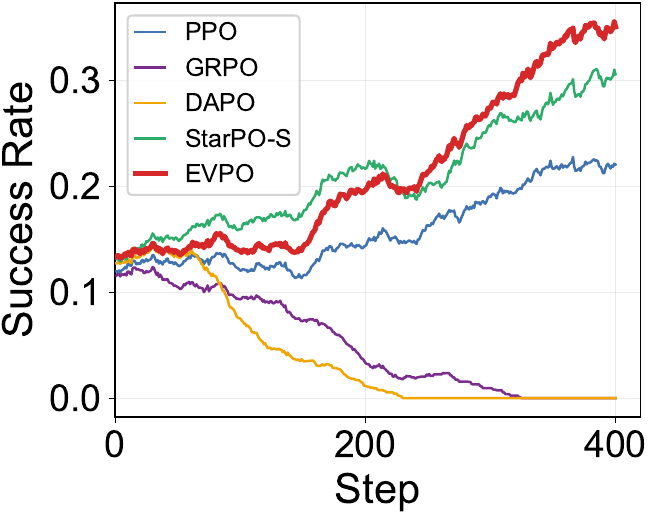}
        \caption{Sokoban (Train)}
    \end{subfigure}
    \hfill
    \begin{subfigure}[b]{0.24\textwidth}
        \centering
        \includegraphics[width=\textwidth]{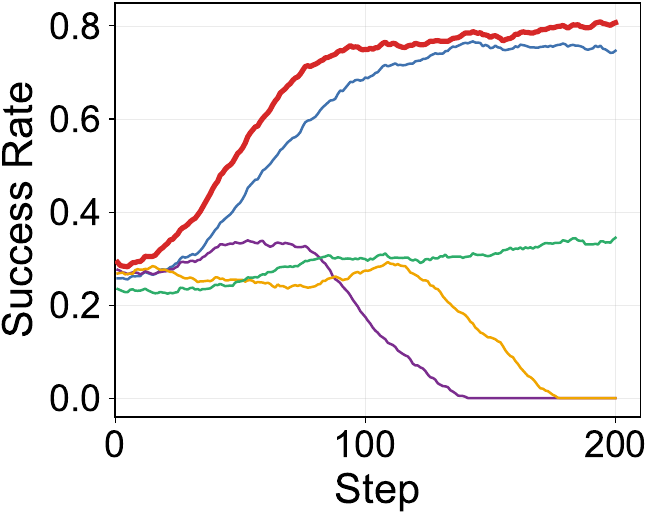}
        \caption{FrozenLake (Train)}
    \end{subfigure}
    \hfill
    \begin{subfigure}[b]{0.24\textwidth}
        \centering
        \includegraphics[width=\textwidth]{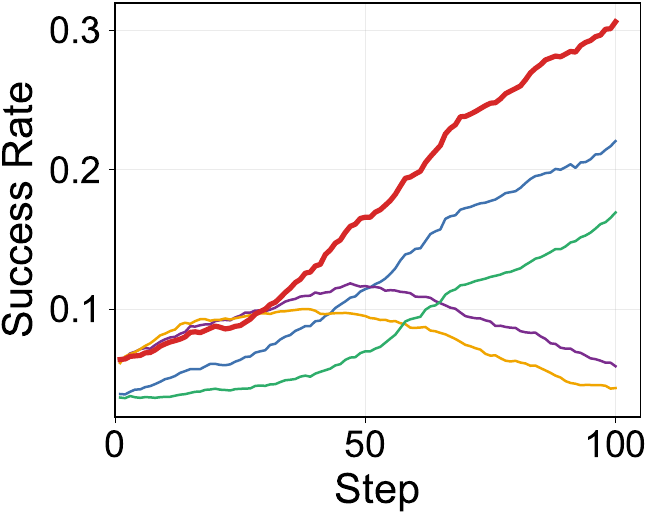}
        \caption{WebShop (Train)}
    \end{subfigure}
    \hfill
    \begin{subfigure}[b]{0.24\textwidth}
        \centering
        \includegraphics[width=\textwidth]{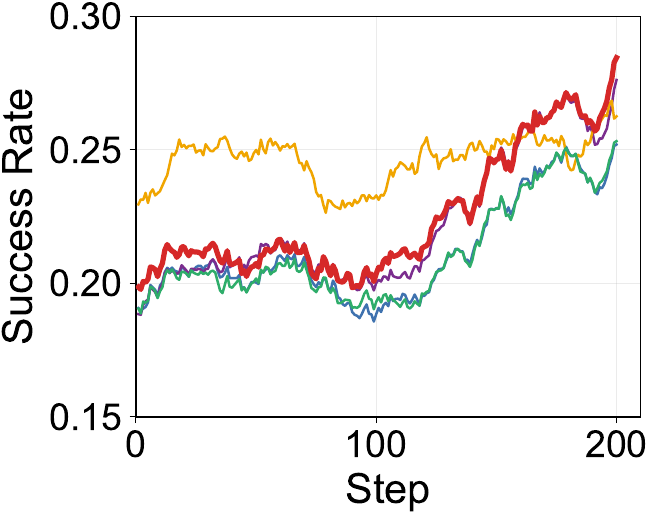}
        \caption{MATH (Train)}
    \end{subfigure}

    \vspace{2mm}

    \begin{subfigure}[b]{0.24\textwidth}
        \centering
        \includegraphics[width=\textwidth]{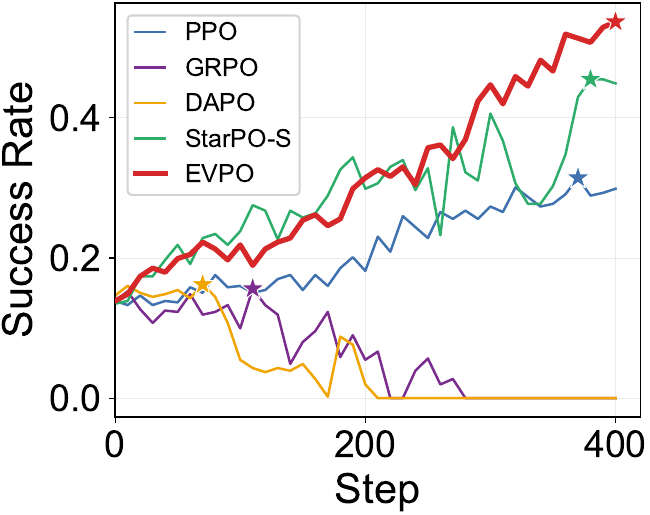}
        \caption{Sokoban (Val)}
    \end{subfigure}
    \hfill
    \begin{subfigure}[b]{0.24\textwidth}
        \centering
        \includegraphics[width=\textwidth]{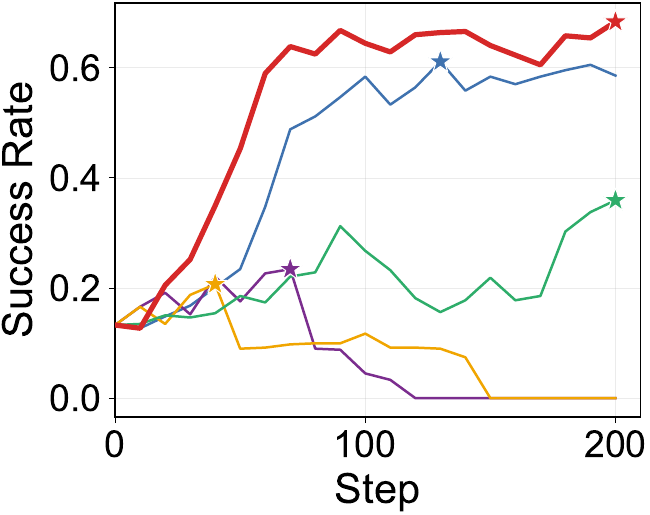}
        \caption{FrozenLake (Val)}
    \end{subfigure}
    \hfill
    \begin{subfigure}[b]{0.24\textwidth}
        \centering
        \includegraphics[width=\textwidth]{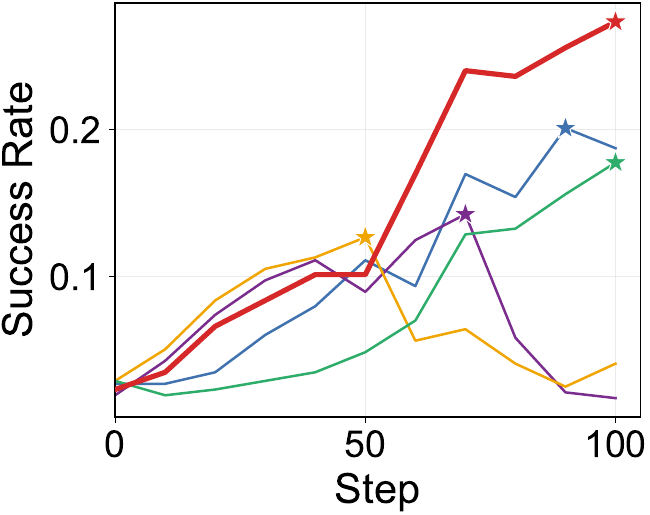}
        \caption{WebShop (Val)}
    \end{subfigure}
    \hfill
    \begin{subfigure}[b]{0.24\textwidth}
        \centering
        \includegraphics[width=\textwidth]{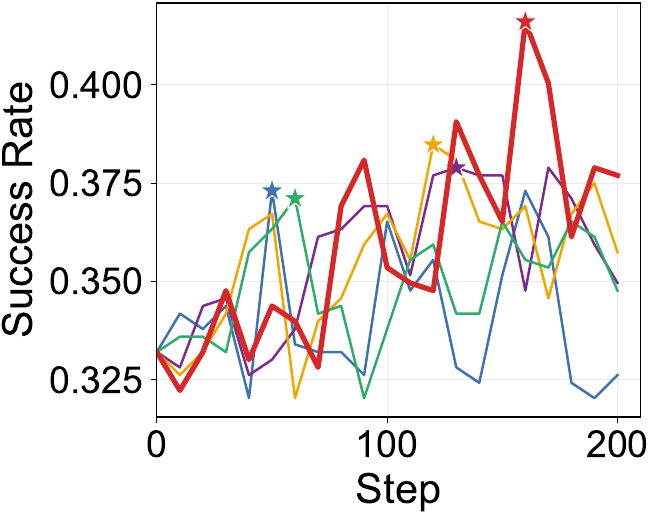}
        \caption{MATH (Val)}
    \end{subfigure}
\caption{Training and validation curves across four tasks. \textbf{Top row:} training success rate (smoothed). \textbf{Bottom row:} validation success rate (unsmoothed; stars mark each method's peak). \method tracks or exceeds the best competing method throughout most of training.}
\label{fig:training_curves}
\end{figure}


\begin{findingBox}{2}{
\method's advantage persists throughout training, not just at convergence.
}
\end{findingBox}

As shown in Figure~\ref{fig:training_curves}, \method leads or matches the strongest baseline at most training stages across all four tasks. This indicates that adaptive switching takes effect continuously---not by recovering from a poor start or catching up late, but by maintaining lower-variance advantage estimates throughout the optimization trajectory. The per-step variance guarantee (Eq.~\ref{eq:evpo_var}) predicts exactly this behavior: since EVPO selects the lower-variance baseline at every step, its cumulative benefit should manifest as a sustained advantage rather than a late-stage correction.

\subsection{Gating Behavior Analysis}
\label{sec:gating}

\begin{figure}[htbp]
\centering
    \begin{subfigure}[b]{0.24\textwidth}
        \centering
        \includegraphics[width=\textwidth]{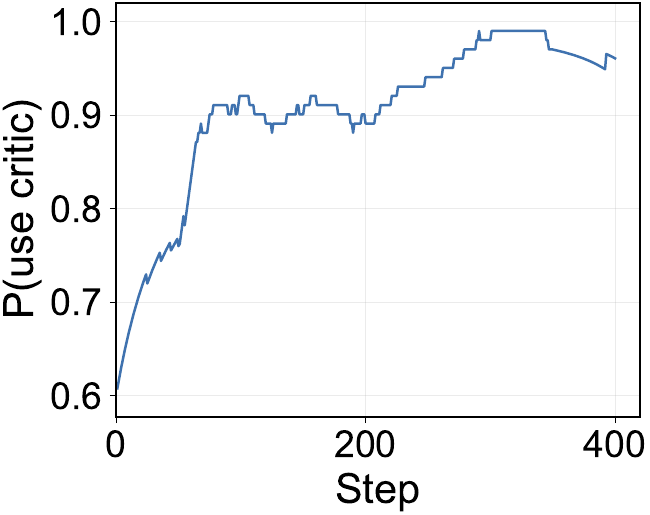}
        \caption{Sokoban}
    \end{subfigure}
    \hfill
    \begin{subfigure}[b]{0.24\textwidth}
        \centering
        \includegraphics[width=\textwidth]{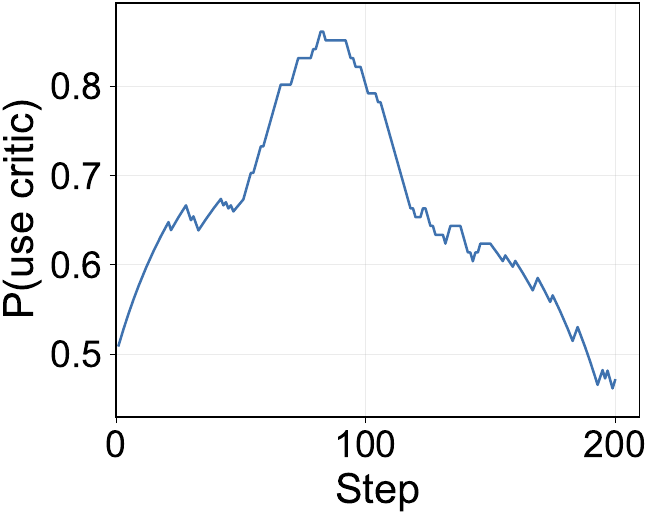}
        \caption{FrozenLake}
    \end{subfigure}
    \hfill
    \begin{subfigure}[b]{0.24\textwidth}
        \centering
        \includegraphics[width=\textwidth]{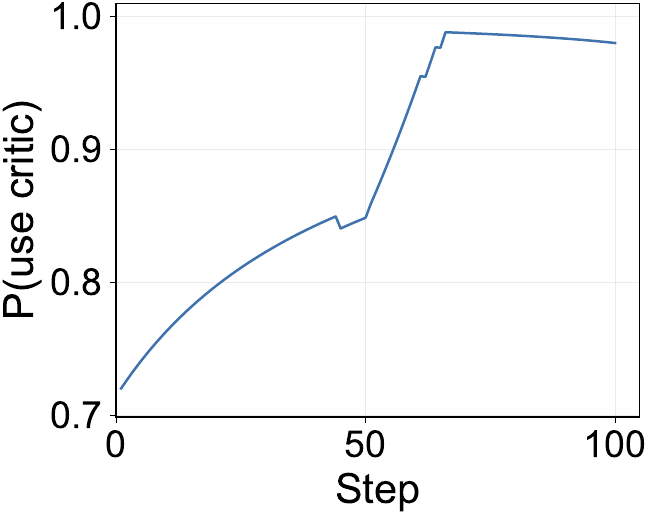}
        \caption{WebShop}
    \end{subfigure}
    \hfill
    \begin{subfigure}[b]{0.24\textwidth}
        \centering
        \includegraphics[width=\textwidth]{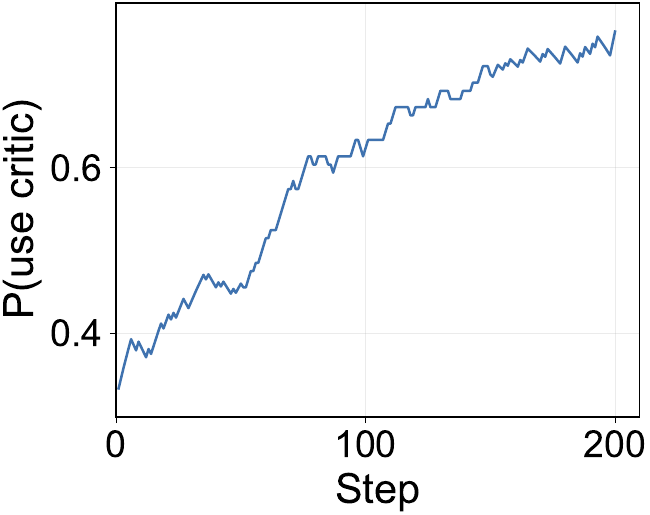}
        \caption{MATH}
    \end{subfigure}
    \caption{
    Gating trigger probability ($\widehat{\ev}_{\mathcal{B}} > 0$, i.e., selection of critic over batch mean) across training.
    }
    \label{fig:gate_curves}
\end{figure}


\begin{findingBox}{3}{
In most tasks, gating frequency decreases monotonically over training, reflecting critic maturation from $P_A > P_B$ to $P_A < P_B$.
}
\end{findingBox}
Figure~\ref{fig:gate_curves} tracks the critic-selection probability over training. Across Sokoban, WebShop, and MATH, the batch-mean mode is triggered frequently in early training and progressively less as training continues, consistent with the expectation that $P_A$ decreases as the critic improves. A notable exception is FrozenLake, where gating frequency reverses around step 80, coinciding with training stagnation (Figure~\ref{fig:training_curves}(b)). This suggests that when the policy enters a difficult region, critic quality can temporarily regress, and the adaptive gating responds by retreating to the batch-mean baseline.

\subsection{EV Threshold Sensitivity}
\label{sec:threshold}

\begin{figure}[htbp]
\centering
    \begin{subfigure}[b]{0.24\textwidth}
        \centering
        \includegraphics[width=\textwidth]{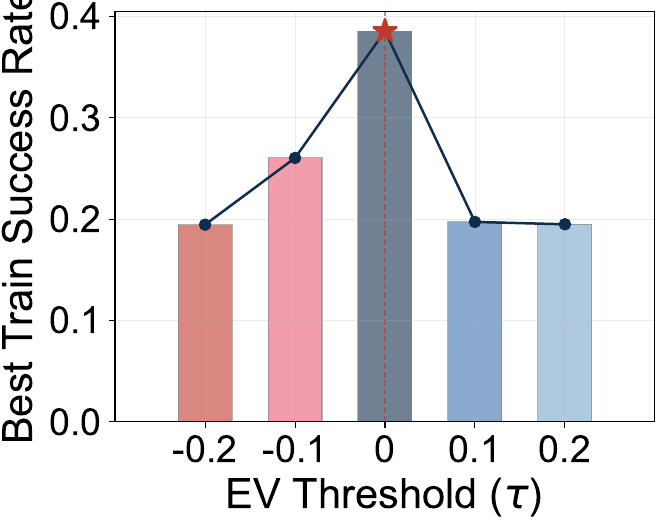}
        \caption{Sokoban (Train)}
    \end{subfigure}
    \hfill
    \begin{subfigure}[b]{0.24\textwidth}
        \centering
        \includegraphics[width=\textwidth]{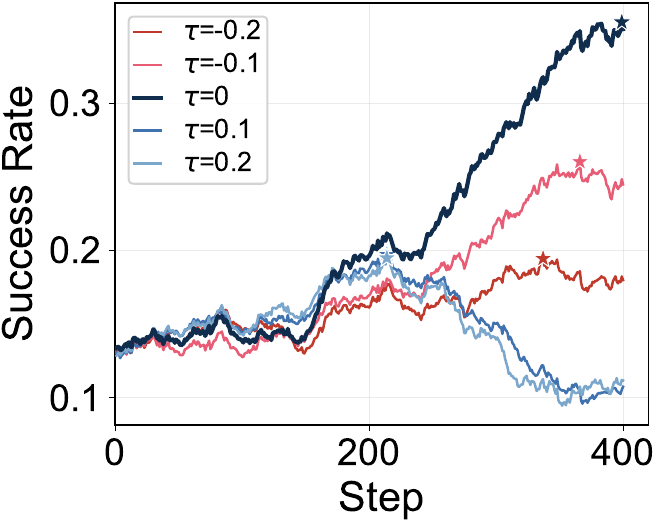}
        \caption{Sokoban (Train)}
    \end{subfigure}
    \hfill
    \begin{subfigure}[b]{0.24\textwidth}
        \centering
        \includegraphics[width=\textwidth]{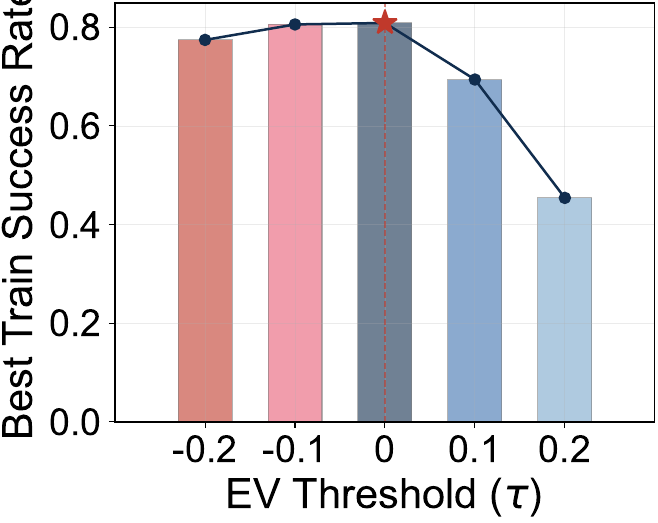}
        \caption{FrozenLake (Train)}
    \end{subfigure}
    \hfill
    \begin{subfigure}[b]{0.24\textwidth}
        \centering
        \includegraphics[width=\textwidth]{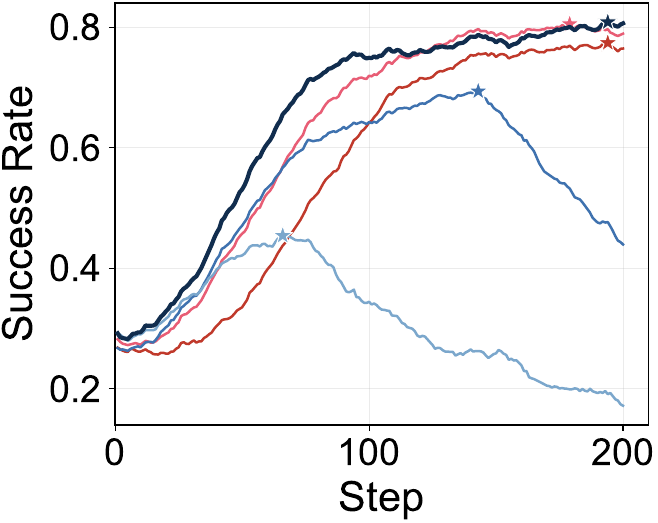}
        \caption{FrozenLake (Train)}
    \end{subfigure}

    \vspace{2mm}

    \begin{subfigure}[b]{0.24\textwidth}
        \centering
        \includegraphics[width=\textwidth]{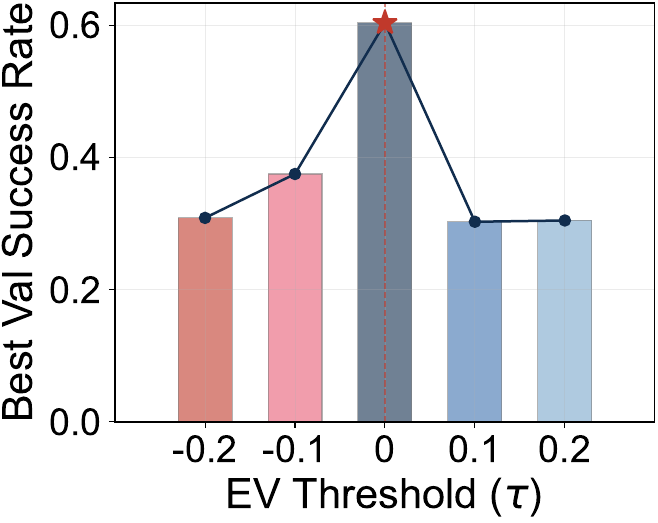}
        \caption{Sokoban (Val)}
    \end{subfigure}
    \hfill
    \begin{subfigure}[b]{0.24\textwidth}
        \centering
        \includegraphics[width=\textwidth]{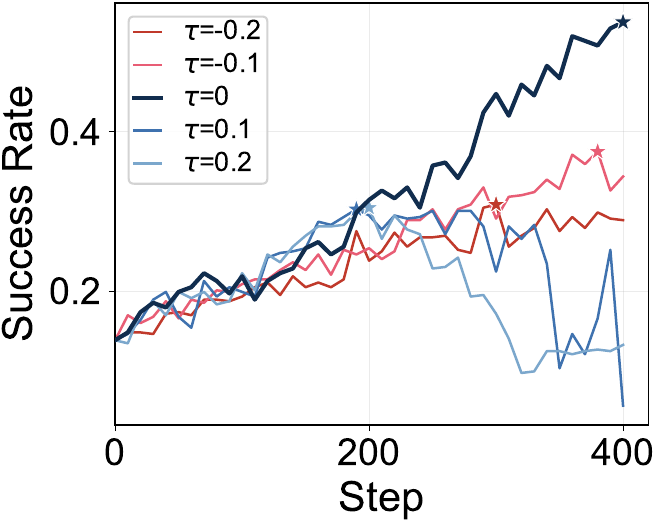}
        \caption{Sokoban (Val)}
    \end{subfigure}
    \hfill
    \begin{subfigure}[b]{0.24\textwidth}
        \centering
        \includegraphics[width=\textwidth]{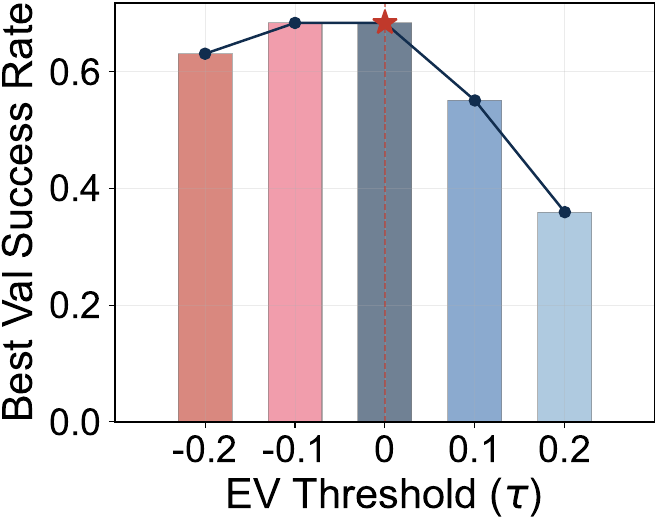}
        \caption{FrozenLake (Val)}
    \end{subfigure}
    \hfill
    \begin{subfigure}[b]{0.24\textwidth}
        \centering
        \includegraphics[width=\textwidth]{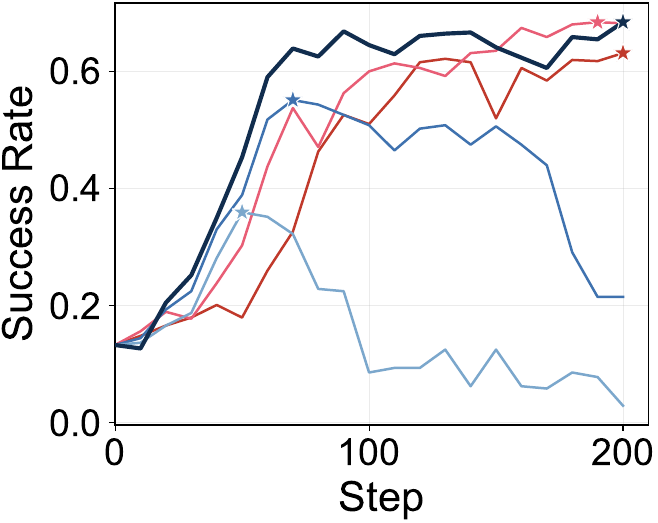}
        \caption{FrozenLake (Val)}
    \end{subfigure}
\caption{Threshold sensitivity analysis. \textbf{Top row:} training metrics (smoothed); \textbf{bottom row:} validation metrics (unsmoothed). \textbf{(a, c, e, g)}~Best success rate as a function of the EV switching threshold; stars mark $\tau{=}0$. \textbf{(b, d, f, h)}~Full curves under each threshold setting. Performance peaks at $\tau{=}0$ on both tasks.}
\label{fig:threshold}
\end{figure}




\begin{findingBox}{4}{
    Performance peaks at $\ev = 0$ and degrades in either direction, with positive and negative deviations inducing distinct failure modes.
}
\end{findingBox}
We vary the switching threshold across $\{-0.2, -0.1, 0, 0.1, 0.2\}$ while keeping all other configurations fixed.
As shown in Figure~\ref{fig:threshold}(a, c, e, g), the theoretically derived boundary $\ev = 0$ (Theorem~\ref{thm:collapse}) is not merely a theoretical convenience but the practically optimal operating point.
The training landscape in Figure~\ref{fig:threshold}(b, d, f, h) reveals an asymmetry in the two deviation directions. A positive threshold under-utilizes the critic: training progresses normally in early stages but collapses later when policy refinement requires state-dependent advantage estimates that the batch mean cannot provide. A negative threshold over-utilizes the critic: the immature critic's noise slows early training, and in later stages, unreliable value predictions on certain state subsets continue to inject variance. These two failure modes reveal that, as we argued theoretically, the zero threshold is precisely the boundary separating the $P_A > P_B$ and $P_A < P_B$ regimes. 


%% file: sections/Related_Work.tex
\section{Related Work}
\label{sec:related}

\paragraph{Reinforcement Learning with Verifiable Rewards for LLMs.}
RL-based LLM post-training has evolved from actor-critic architectures such as PPO~\citep{schulmanProximalPolicyOptimization2017}, which rely on a learned value function for variance reduction via GAE~\citep{schulmanHighDimensionalContinuousControl2018}, toward critic-free alternatives that bypass the value network entirely.
GRPO~\citep{shaoDeepSeekMathPushingLimits2024,guoDeepSeekR1IncentivizesReasoning2025,wangEnhancingLLMbasedSearch2026} replaces the learned baseline with intra-group reward normalization and has become the dominant paradigm; REINFORCE++~\citep{huREINFORCEStabilizingCriticFree2025} further stabilizes this family via global batch normalization, and similar estimators like RLOO~\citep{ahmadianBackBasicsRevisiting2024} and DAPO~\citep{yuDAPOOpenSourceLLM2025a} further demonstrate that a critic is often dispensable for single-turn, verifiable tasks.

Yet critic-free methods face two systematic challenges: group-relative advantages introduce statistical bias, systematically underestimating hard prompts and overestimating easy ones~\citep{jainUnderstandingOptimizationLandscape2025,linMMDocR1TrainingAgents2026}, and uniform trajectory-level credit assignment degrades severely in long-horizon settings lacking early termination~\citep{oliveiraLearningCriticsRevisiting2025,dongAgenticReinforcedPolicy2025a}, compelling multi-turn frameworks such as RAGEN~\citep{wangRAGENUnderstandingSelfEvolution2025} to reintroduce the critic for stable training.
All existing approaches resolve this tension between critic-free simplicity and critic-based credit assignment~\citep{liUnifyingPPO2025} through a static, a priori architectural commitment that either always uses or always discards the critic, regardless of its current predictive quality.

\paragraph{Adaptive Baselines and Explained Variance.}
Reducing policy-gradient variance without detrimental bias is a foundational RL challenge~\citep{williamsSimpleStatisticalGradientfollowing1992,suttonPolicyGradientMethods1999}. 
State-dependent baselines are theoretically unbiased but rely on the assumption that the value network is accurate; in LLM post-training, the vast state space and sparse rewards frequently violate this assumption, causing the baseline subtraction to inject more noise than it removes.
Prior work addresses this through two complementary strategies: redesigning the baseline structure itself~\citep{anschelAveragedDQNVarianceReduction2017b}, or using critic quality as a signal to adapt hyperparameters such as $\lambda$ in GAE~\citep{NaemprPPOwithadaptiveGAE} or to filter uninformative transitions~\citep{flet-berliacOnlyRelevantInformation2019}.

However, these methods treat critic quality as a signal to tune hyperparameters or curate data while preserving the actor-critic architecture.
\method departs from both strategies by elevating EV from a passive diagnostic to a architecture selection criterion that determines \emph{whether} the critic is used at all, switching to critic-free estimation when the value network becomes mathematically harmful.

%% file: sections/Conclusion.tex
\section{Conclusion}
\label{sec:conclusion}

We presented \method, an adaptive RL method for LLM post-training that uses explained variance to switch between critic-based and batch-mean advantage estimation at each training step. By casting baseline selection as a Kalman filtering problem, we proved that the sign of EV is the exact boundary separating the variance-reducing from the variance-inflating critic regime (Theorem~\ref{thm:collapse}), and that \method's per-step switching yields advantage variance no greater than the better of PPO and GRPO at every step.

Across four tasks spanning multi-turn planning, agentic interaction, and mathematical reasoning, \method consistently outperforms both fixed baselines, regardless of which one is stronger on a given task. The gating analysis confirms that \method's adaptation tracks critic maturation over training, and the threshold sensitivity study validates that the theoretically derived zero boundary is empirically optimal.


%% file: sections/Limitations.tex
\section*{Limitations}
\label{sec:limitations}

\paragraph{Population vs.\ batch EV.}
Theorem~\ref{thm:collapse} characterizes the collapse boundary in terms of population EV, whereas \method switches on batch EV computed from finite samples. We do not provide finite-sample concentration bounds for the switching rule, though the threshold sensitivity analysis (Section~\ref{sec:threshold}) suggests that zero is a robust practical threshold.

\paragraph{Resource overhead.}
\method maintains a critic network throughout training, including iterations where the batch-mean mode is selected, and therefore retains the memory footprint of PPO rather than matching fully critic-free methods.

\paragraph{Task coverage.}
Our evaluation spans four tasks across three domains. Tasks with dense intermediate rewards, learned reward models, or substantially different rollout lengths may exhibit different EV dynamics.

%% file: sections/app_preliminary.tex
\section{Extended Proofs and Derivations}
\label{app:proofs}
Detailed proofs and derivations are provided in this appendix.

\subsection{Detailed Kalman Filtering Analogy}
\label{app:kalman_details}

We provide here the full state estimation analogy and derivation that motivates the EV-based switching criterion introduced in Section~\ref{sec:kalman}.

\paragraph{State estimation analogy.}
We cast baseline construction as a scalar state estimation problem.
The hidden state to be estimated is the true value function $x = V^\pi(s)$.
For each sample in a training batch, the observed return decomposes as
\begin{equation}
  G = V^\pi(s) + \epsilon, \quad \epsilon \sim (0, R), \quad R \triangleq \E[\Var(G \mid s)],
  \label{eq:return_decomp}
\end{equation}
where $R$ is the irreducible \emph{sampling noise} arising from the stochasticity of future actions under $\pi$.
Two competing estimators of $V^\pi(s)$ are available at each training step:

\begin{itemize}
  \item \textbf{Model-driven prediction (Critic).}
    $\hat{x}_A = \hat{V}_\phi(s) = V^\pi(s) + \delta(s)$,
    where $\delta(s)$ denotes the critic's estimation error with variance
    $P_A \triangleq \Var(\delta)$.
    This parallels the \emph{state prediction} in a Kalman filter:
    it exploits learned state features but carries modelling noise whose magnitude depends on the critic's current training quality.

  \item \textbf{Evidence-driven observation (Batch mean).}
    $\hat{x}_B = \bar{G}_{\mathcal{B}}$, the empirical mean return over the batch, which approximates $\E[V^\pi(s)]$.
    Its error relative to any specific state $s$ has variance
    $P_B \triangleq \Var_s(V^\pi(s))$,
    paralleling the \emph{observation noise} in a Kalman filter:
    it reflects how much true state values vary within the batch, and is large when the batch covers heterogeneous states.
\end{itemize}

\noindent
The sampling noise $R$ affects both estimators symmetrically and controls the \emph{sensitivity} of the EV diagnostic without altering its \emph{sign}.

\paragraph{Optimal fusion and the Kalman gain.}
The Kalman filter combines the two estimators via a linear combination
$b(s) = (1 - K)\,\hat{x}_A + K\,\hat{x}_B$
that minimizes the mean squared error $\E[(b(s) - V^\pi(s))^2]$.
The optimal \emph{Kalman gain} is:
\begin{equation}
  K^* = \frac{P_A}{P_A + P_B}.
  \label{eq:kalman_gain_app}
\end{equation}
When $K^* < \tfrac{1}{2}$ (i.e., $P_A < P_B$), the critic is more reliable than the batch mean and should receive higher weight;
when $K^* \ge \tfrac{1}{2}$ (i.e., $P_A \ge P_B$), the critic's estimation noise dominates the state signal it provides, and the baseline should retreat toward the batch mean.

\paragraph{Connection to explained variance.}
Substituting the variance decompositions from Eq.~\eqref{eq:return_decomp} into the definition of EV yields:
\begin{equation}
  \ev = \frac{P_B - P_A}{P_B + R}.
\end{equation}
Since the denominator is always positive, the sign of EV is determined solely by $P_B - P_A$:
\begin{itemize}
  \item $\ev > 0 \;\Longleftrightarrow\; P_A < P_B \;\Longleftrightarrow\; K^* < \tfrac{1}{2}$:
    the critic outperforms the batch mean, corresponding to the regime where PPO's state-dependent baseline is advantageous;
  \item $\ev \le 0 \;\Longleftrightarrow\; P_A \ge P_B \;\Longleftrightarrow\; K^* \ge \tfrac{1}{2}$:
    the critic's noise exceeds its signal, corresponding to the regime where GRPO's constant baseline is preferable.
\end{itemize}

\subsection{GAE Derivation under RLVR Conditions}
\label{app:gae_derivation}

Proximal Policy Optimization~\citep{schulmanProximalPolicyOptimization2017} estimates advantages via Generalized Advantage Estimation (GAE)~\citep{schulmanHighDimensionalContinuousControl2018}:
\begin{equation}
  \hat{A}_t^{\mathrm{GAE}} = \sum_{l=0}^{T-1-t} (\gamma \lambda)^l \delta_{t+l}, \quad
  \delta_t = r_t + \gamma \hat{V}_\phi(s_{t+1}) - \hat{V}_\phi(s_t),
  \label{eq:gae}
\end{equation}
where $\hat{V}_\phi$ is a learned critic and $\lambda \in [0,1]$ controls the bias--variance trade-off.
The policy is updated by maximizing a clipped surrogate objective that constrains the ratio $\pi_\theta(a_t|s_t) / \pi_{\theta_{\text{old}}}(a_t|s_t)$.

Under RLVR conditions ($\gamma = \lambda = 1$, terminal-only reward $r_T = \mathcal{R}(x,y)$, $r_t = 0$ for $t < T$), GAE reduces to:
\begin{equation}
  \hat{A}_t^{\mathrm{PPO}} = \sum_{l=t}^{T-1} \bigl(r_l + \hat{V}_\phi(s_{l+1}) - \hat{V}_\phi(s_l)\bigr) = G - \hat{V}_\phi(s_t),
\end{equation}
where the telescoping sum cancels all intermediate critic terms (with the terminal state value defined as 0), leaving the advantage as the deviation of the observed return from the critic's state-dependent prediction.

%% file: sections/app_exp_details.tex
\section{Tasks}
\label{app:exp_details}
\label{app:tasks}

We evaluate on four tasks spanning multi-turn grid-world control, interactive
web agents, and single-turn mathematical reasoning.
All training code is based on the \texttt{RAGEN2} framework~\citep{wangRAGENUnderstandingSelfEvolution2025},
which wraps \texttt{verl} for PPO-style LLM post-training and provides
multi-turn agent environments.

\paragraph{Sokoban.}
We use the \texttt{CoordSokoban} variant of the RAGEN2 Sokoban environment~\citep{junghannsSokobanEnhancingGeneral2001}.
Each episode instantiates a $6\times6$ grid with one box and one target,
procedurally generated from an environment seed.
The observation at each step is a symbol grid together with zero-indexed
coordinates of the player, the box, and the target.
The action space is $\{\textsc{Up},\textsc{Down},\textsc{Left},\textsc{Right}\}$;
the policy is prompted to emit an action sequence of length up to two per turn,
separated by ``\texttt{||}'' and enclosed in \texttt{<answer>\ldots</answer>} tags.
A trajectory consists of up to $10$ executed actions across up to $5$ turns.
The reward is terminal and binary: $+1$ if all boxes end on targets, $0$ otherwise,
with an additional format penalty of $-0.1$ for malformed outputs.

\paragraph{FrozenLake.}
We use the \texttt{CoordFrozenLake} variant wrapping the Gymnasium
\texttt{FrozenLake-v1} environment~\citep{PDFFrozenLake}.
The map is a $4\times4$ grid with randomly placed holes, the start tile, and
the goal tile; the map layout is resampled per environment seed.
Transitions are stochastic (\texttt{is\_slippery=True}): the commanded action
succeeds with probability $1/3$, and with probability $2/3$ the agent slips
perpendicular to the intended direction. The observation is again a symbol grid
plus coordinates of start, goal, player, and all holes.
The action space is $\{\textsc{Left},\textsc{Down},\textsc{Right},\textsc{Up}\}$,
with up to $10$ actions per trajectory across up to $5$ turns.
The reward is terminal and binary: $+1$ if the player reaches the goal,
$0$ if the player falls into a hole or exhausts the action budget;
a format penalty of $-0.1$ is applied for malformed outputs.

\paragraph{WebShop.}
WebShop~\citep{yaoWebShopScalableRealWorld2022} is an interactive web-navigation task in which the
agent must locate and purchase a product matching a natural-language instruction.
We use the \texttt{webshop-minimal} fork with the \emph{small} product catalogue
shipped with RAGEN2.
At each turn the agent observes the current page (search box, product list,
or product detail) together with the list of clickable elements, and produces
exactly one action drawn from \texttt{search[\textit{keywords}]} or
\texttt{click[\textit{element}]}, again wrapped in \texttt{<answer>\ldots</answer>}.
A trajectory consists of up to $9$ actions across up to $9$ turns.
The reward is the built-in WebShop matching score in $[0,1]$,
delivered only at the terminal \texttt{click[buy now]}.

\paragraph{MATH.}
We use DAPO-Math-17k~\citep{yuDAPOOpenSourceLLM2025a}
(17{,}267 competition-level problems) for \emph{both} training and validation:
the full training set is used for RL updates, and a \emph{fixed subset of 32
problems}, held out from training, is used for evaluation.
Each problem is a single-turn reasoning task (\texttt{max\_turn=1}):
the policy is prompted with the problem statement and must emit a
\texttt{<think>\ldots</think>} block followed by the final numerical answer
inside \texttt{<answer>\ldots</answer>}.
The reward is terminal and binary: $+1$ if the final answer matches the ground
truth (after normalization), $0$ otherwise.

\section{Baselines and Hyperparameters}
\label{app:baselines}
\label{app:hyper}

We compare EVPO against a Base-LLM reference and four RL baselines that span
both critic-based and critic-free families.

\begin{itemize}
  \item \textbf{Base LLM.} Instruction-tuned Qwen2.5 without any RL post-training,
    prompted zero-shot with the same task template as the RL runs.
  \item \textbf{PPO}~\citep{schulmanProximalPolicyOptimization2017}.
    Standard clipped-surrogate PPO with GAE advantages. Under our RLVR setting
    ($\gamma{=}\lambda{=}1$, terminal-only reward) this reduces to the PPO advantage
    of Eq.~\eqref{eq:ppo_adv} with a learned, state-dependent critic.
  \item \textbf{StarPO-S}~\citep{wangRAGENUnderstandingSelfEvolution2025}.
    A multi-turn PPO variant introduced by RAGEN that filters the lowest-variance
    rollouts before each update
    (\texttt{rollout\_filter\_ratio}$=0.25$,
    \texttt{rollout\_filter\_metric}$=$reward\_variance) to stabilize long-horizon
    agentic training.
  \item \textbf{GRPO}~\citep{shaoDeepSeekMathPushingLimits2024}.
    Critic-free; the baseline is the response-level batch mean, normalized by the
    batch standard deviation.
  \item \textbf{DAPO}~\citep{yuDAPOOpenSourceLLM2025a}.
    A modern critic-free variant built on GRPO with (i) dynamic sampling that
    discards groups whose rewards are all identical and resamples up to
    $\texttt{max\_num\_gen\_batches}{=}5$ times until enough informative groups
    are collected; (ii) \emph{clip-higher} asymmetry
    $(\varepsilon_\text{low},\varepsilon_\text{high})=(0.2,0.28)$;
    (iii) token-level loss aggregation
    (\texttt{loss\_agg\_mode}$=$token-mean); and
    (iv) no std normalization of advantages and no in-reward KL term.
\end{itemize}

All five methods (PPO, StarPO-S, GRPO, DAPO, EVPO) share the same rollout stack,
tokenizer, and environment code, and differ only in the advantage
estimator and the listed hyperparameters. EVPO itself uses PPO-style training
(critic and actor with clipped surrogate) and only changes the baseline subtracted
from the return, as prescribed by the EV-gating rule (Eq.~\eqref{eq:evpo_rule}).

Table~\ref{tab:app-hyper} lists the hyperparameters that differ across
methods or tasks. Learning rates are tuned per task on the
Base-LLM vs.\ PPO gap on a single seed and then \emph{kept identical across
methods on the same task} for fair comparison, except where a differing
value is itself a defining feature of the algorithm (see the caption of
Table~\ref{tab:app-hyper}).

\begin{table}[h]
  \centering
  \small
  \caption{Per-task / per-method hyperparameter overrides.
  $\varepsilon_\text{lo}/\varepsilon_\text{hi}$ are the PPO-style clip
  thresholds. Where DAPO and StarPO-S show different values from the
  other methods on the same task (e.g., DAPO's clip-higher asymmetry,
  or the StarPO-S Sokoban configuration that preserves the original
  RAGEN recipe), those settings are defining features of the respective
  algorithms rather than tuning choices, and we therefore do not force
  them to match the other methods.}
  \label{tab:app-hyper}
  \begin{tabular}{@{}llccccc@{}}
    \toprule
    Task & Method & lr$_\text{actor}$ & lr$_\text{critic}$
         & $\varepsilon_\text{lo}/\varepsilon_\text{hi}$
         & KL coef. & Steps \\
    \midrule
    \multirow{5}{*}{Sokoban ($6{\times}6$)}
      & PPO       & $4\mathrm{e}{-}6$ & $1\mathrm{e}{-}5$ & $0.20/0.28$ & $0$ & $500$ \\
      & StarPO-S  & $1\mathrm{e}{-}6$ & $1\mathrm{e}{-}5$ & $0.20/0.28$ & $0$ & $400$ \\
      & GRPO      & $4\mathrm{e}{-}6$ & -- & $0.20/0.28$ & $0$ & $400$ \\
      & DAPO      & $4\mathrm{e}{-}6$ & -- & $0.20/0.28$ & $0$ & $400$ \\
      & \textbf{EVPO} & $4\mathrm{e}{-}6$ & $1\mathrm{e}{-}5$ & $0.20/0.28$ & $0$ & $500$ \\
    \midrule
    \multirow{5}{*}{FrozenLake ($4{\times}4$)}
      & PPO       & $1\mathrm{e}{-}6$ & $1\mathrm{e}{-}5$ & $0.20/0.28$ & $0$ & $200$ \\
      & StarPO-S  & $1\mathrm{e}{-}6$ & $1\mathrm{e}{-}5$ & $0.20/0.28$ & $0$ & $200$ \\
      & GRPO      & $1\mathrm{e}{-}6$ & -- & $0.20/0.28$ & $0$ & $200$ \\
      & DAPO      & $1\mathrm{e}{-}6$ & -- & $0.20/0.28$ & $0$ & $200$ \\
      & \textbf{EVPO} & $1\mathrm{e}{-}6$ & $1\mathrm{e}{-}5$ & $0.20/0.28$ & $0$ & $200$ \\
    \midrule
    \multirow{5}{*}{WebShop}
      & PPO       & $5\mathrm{e}{-}7$ & $5\mathrm{e}{-}6$ & $0.20/0.20$ & $10^{-2}$ & $100$ \\
      & StarPO-S  & $1\mathrm{e}{-}6$ & $1\mathrm{e}{-}5$ & $0.20/0.28$ & $0$ & $100$ \\
      & GRPO      & $5\mathrm{e}{-}7$ & -- & $0.20/0.20$ & $10^{-2}$ & $100$ \\
      & DAPO      & $5\mathrm{e}{-}7$ & -- & $0.20/0.28$ & $10^{-2}$ & $100$ \\
      & \textbf{EVPO} & $5\mathrm{e}{-}7$ & $5\mathrm{e}{-}6$ & $0.20/0.20$ & $10^{-2}$ & $100$ \\
    \midrule
    \multirow{5}{*}{MATH}
      & PPO       & $2\mathrm{e}{-}6$ & $1\mathrm{e}{-}5$ & $0.20/0.20$ & $0$ & $200$ \\
      & StarPO-S  & $2\mathrm{e}{-}6$ & $1\mathrm{e}{-}5$ & $0.20/0.28$ & $0$ & $200$ \\
      & GRPO      & $2\mathrm{e}{-}6$ & -- & $0.20/0.20$ & $0$ & $200$ \\
      & DAPO      & $2\mathrm{e}{-}6$ & -- & $0.20/0.28$ & $0$ & $200$ \\
      & \textbf{EVPO} & $2\mathrm{e}{-}6$ & $1\mathrm{e}{-}5$ & $0.20/0.20$ & $0$ & $200$ \\
    \bottomrule
  \end{tabular}
\end{table}

\section{Details of the Preliminary Experiments}
\label{app:prelim}

This section gives the exact protocols for the two preliminary studies that
motivate EVPO in Section~\ref{sec:prelim}.

\subsection{PPO on Sokoban: EV, Gradient Norm, Success Rate (Figure~\ref{fig:obs1})}
\label{app:prelim-ppo}
\label{app:preliminary_exp1}

We train a PPO agent on Sokoban using the same setting as the Sokoban
PPO main experiment (Sokoban PPO row of Table~\ref{tab:app-hyper}).
At every step we record, over the same batch of $M=16\times8=128$ rollouts used
for the update:
(i) the batch-level EV $\widehat{\ev}_\mathcal{B}$ defined in Eq.~\eqref{eq:ev_batch};
(ii) the pre-clipping gradient norm of the actor;
(iii) the training success rate (fraction of rollouts with terminal reward $=1$).
Shaded regions in Figure~\ref{fig:obs1}(a--c) highlight the first
$\approx 150$ steps where $\widehat{\ev}_\mathcal{B}$ remains near or below $0$.

\paragraph{Noise-injection experiment (Figure~\ref{fig:obs1d}).}
We first train a PPO agent on FrozenLake to convergence under the same
hyperparameters as the FrozenLake row of Table~\ref{tab:app-hyper} for $200$
steps, then continue training with i.i.d.\ Gaussian noise added to every
critic output: $\hat V_\phi(s) \mapsto \hat V_\phi(s) + \xi$ with
$\xi \sim \mathcal{N}(0,\sigma_\delta^2)$. We sweep
$\sigma_\delta \in \{0.3,\,1,\,3,\,10\}$, corresponding to noise magnitudes
from well below to well above the mean absolute state value
($\approx 0.5$). Each noise level is run for an additional $150$ training
steps, starting from the same converged checkpoint. We report the smoothed
validation success rate; Fig.~\ref{fig:obs1d} shows monotone degradation
with complete collapse at $\sigma_\delta{=}10$, precisely the regime where
$\widehat{\ev}_\mathcal{B}$ is driven far below zero.

\subsection{Cold-start and Critic Warmup (Figure~\ref{fig:obs2})}
\label{app:prelim-warmup}
\label{app:preliminary_exp2}

We run two interventions on top of the Sokoban and FrozenLake PPO setups.

\paragraph{Cold-start.}
During the first $K_\text{cold}$ training steps we replace every critic
output in the advantage computation by the batch mean,
$\hat A_t^{\text{cold-start}} = G - \bar G_\mathcal{B}$;
after step $K_\text{cold}$ we resume standard PPO with the original critic.
We set $K_\text{cold}=50$ on Sokoban and $K_\text{cold}=25$ on FrozenLake,
chosen to cover the EV-negative phase of Fig.~\ref{fig:obs1a}.
The critic itself is always trained with its usual MSE loss so that by step
$K_\text{cold}$ it has had time to warm up.

\paragraph{Critic warmup.}
Before the actor receives any gradient, we train the critic alone for
$K_\text{warm}=200$ steps using the same rollouts as full PPO but with
$\nabla_\theta \mathcal{L}_\text{actor}$ set to zero. After step
$K_\text{warm}$ we switch on actor updates and continue with standard PPO.

In both interventions all other hyperparameters match the PPO rows of
Table~\ref{tab:app-hyper}. Figure~\ref{fig:obs2} compares each
intervention against the unmodified PPO baseline under the same seed and
hardware; the consistent improvements confirm that it is specifically the
suppression of critic-driven variance (rather than any secondary effect
of the interventions) that benefits the policy.